\documentclass[letterpaper, 10 pt, conference]{ieeeconf}  

\IEEEoverridecommandlockouts                              

\usepackage[utf8]{inputenc}
\usepackage{amsmath}
\usepackage{cleveref}
\usepackage{caption}
\usepackage{subcaption}
\usepackage{graphicx}

\title{\LARGE \bf
Robotic Sculpting\\
with Collision-free Motion Planning 
in Voxel Space
}

\author{Abhinav Jain$^{*}$, Seth Hutchinson$^{*}$, and Frank Dellaert$^{*}$ 
\thanks{$^{*}$All three authors are with the Georgia Institute of Technology, Atlanta, GA 30332, USA
        {\tt\small (jain|seth|fd27)@gatech.edu}}%
}

\begin{document}

\maketitle

\thispagestyle{empty}
\pagestyle{empty}

\begin{abstract}

    In this paper, we explore the task of robot sculpting. We propose a search based planning algorithm to solve the problem of sculpting by material removal with a multi-axis manipulator. We generate collision free trajectories for a manipulator using best-first search in voxel space. We also show significant speedup of our algorithm by using octrees to decompose the voxel space. We demonstrate our algorithm on a multi-axis manipulator in simulation by sculpting Michelangelo's Statue of David, evaluate certain metrics of our algorithm and discuss future goals for the project.

\end{abstract}

\section{Introduction}
\label{sec:introduction}


Sculpting is the act of generating 3D structures through addition or removal of material. In this paper, we address the problem of sculpting with robotic manipulators by removing material. A robot that can sculpt a 3D model from a given material with no human guidance can have diverse applications in fields such as art, manufacturing and medicine. It could be used to create original sculptures as well as replicas of classical sculptures. Robots could also be used to sculpt intricate ornamentations on building facades, which is generally prohibitively expensive due to the amount of highly skilled labour required. Another very important application is in rapid prototyping of parts. It is often assumed that addition of material (as with 3D printing) is a better solution for rapid prototyping. However, with a fast and robust robotic sculpting process, it may be possible to prototype parts by removal of material with minimal human input, allowing the use of stronger materials.


The key unsolved challenge in robotic sculpting is generating optimal collision free trajectories in a dynamic environment, which will enable a robot to sculpt a given 3D model. We address this problem using a search algorithm that sequentially generates collision-free trajectories, ensuring that the trajectories are optimal, and that the dynamic characteristic of the environment is taken into account. We have simplified the task by discretizing the material and the model into a voxel space. While a voxel representation of 3D models may not be feasible for real world applications of sculpting, it allows us to establish a proof of concept of our algorithm, as well as sets the groundwork for future research into applying the search using free-form surfaces.


Not much work has been done in the robotic community  on the topic of robot sculpture.~\cite{xuejuan2007robot} and~\cite{lei20083d} proposed a full sculpture robot system. In their work, they showcased a topography sculptor, which, given a topological map, is able to sculpt the terrain from a block of material. They generated NURBS splines along one direction of the map. The splines form the trajectories for their robot's end effector. They did not perform any complex collision detection under the assumption that no concave structures would be sculpted, allowing the robot easy access from the top.

Sculpting a 3D model through material removal with a multi-link manipulator is an analogous problem to multi-axis machining, a well studied problem in the machining and CAD domains. While multi-axis machining is considered a solved problem, there are many shortcomings in the general solution that make it unsuitable for our application of autonomous sculpting. The pipeline for multi-axis machining involves a CAM software that generates just a translational and rotational trajectory for just the milling tool, a post processor that generates machine G codes, a processor that performs the inverse kinematics for the multi-axis machine, and a simulator that runs ahead of the actual machine to detect collisions. When a collision is detected, the system returns to the preprocessing state to generate a new toolpath to avoid the detected collision~\cite{lasemi2010recent}. This is an iterative process where a solution is generated, tested and corrected till a final solution is found.

Literature in the CAD domain describes collisions of two general types -- local and global. Local collisions involve collisions of the tool tip with material that is not intended to be removed. Global collisions describe broader collisions of all parts of the machine with any part of the material. Several local and global collision detection methods have been proposed.~\cite{ilushin2005precise} presented a ray-tracing method for global collision detection.~\cite{jun2003optimizing} proposed a configuration space search for global collisions.~\cite{Choi1997} and~\cite{Morishige1997} also explored configuration space collision detection. Similarly, other methods based on surface properties~\cite{chen2005local, Bo2016}, graphics assistance~\cite{Wang2006}, and through simulation~\cite{Lauwers2003a} have been suggested as well.~\cite{tang2014algorithms} outline several such methods. However, the global collision avoidance methods in all these works still limit detection to tooltip and tool holder only, still requiring a simulator to detect collisions and reiterate the preprocessing step.

Sculpting can be seen as a coverage problem where the robot has to cover the entire volume to be removed with its end effector. The general approach for robotic coverage involves decomposing the coverage space into convex cells in the workspace, and then naively generating collision free coverage paths inside those cells~\cite{Choset2001,Choset2000,Atkar2005,breitenmoser2010distributed}. These approaches generally assume a translating robot. Thus, avoiding obstacles in the workspace is adequate. Such approaches are not suitable to be used with multi-link redundant manipulators such as in our application since it is significantly harder to map complex obstacles, even those defined geometrically, into the configuration spaces of redundant manipulators~\cite{zaplana2018novel}. Avoiding obstacles in the workspace of such manipulators is not sufficient to ensure collision free trajectories.

\cite{Hess2012} proposes a coverage solution for manipulators on 3D surfaces. However, while their method is feasible for collision detection in a static environment, it will not work in a dynamic environment such as ours where material is constantly being removed. As our workspace expands with the removal of material, newer, more optimal paths will be available, which~\cite{Hess2012}'s work will not be able to use.


Our work makes the following contributions:

\noindent First, we demonstrate planning for a simple translating robot in a voxel space for material removal to sculpt a given model. 

\noindent Second, we present an octree approach to decompose the larger voxel space into smaller segments. We present a search algorithm to generate a trajectory that cover all the nodes of the octree representation of the model that have to be removed. 

\noindent Third, we present a search algorithm to generate manipulator trajectories for voxel removal inside each block of the octree representation of the model. 

We explain our algorithms in \Cref{sec:description}. We go over experiments and results in simulation in \Cref{sec:results}. We verify our algorithms by executing them on a 3D model of the Statue of David and confirming that collision-free trajectories are efficiently generated. Our primary evaluation metric is time taken for the search.  We discuss the performance of each individual search, as well as the advantages and shortcomings. 
Finally, we discuss possible future steps for this work in \Cref{sec:future}

\section{Description}
\label{sec:description}

The problem we are addressing can be stated as follows:

Given some material of known dimensions, a 3D model of an object that fits within the dimensions of the material, and a robotic manipulator that can remove material at its end effector, compute an optimal collision-free trajectory for the manipulator that will sculpt the 3D model from the material by removing material at the manipulator's end effector.

\subsection{The Voxel Space Approach}
\label{sec:voxel}

\begin{figure}[ht!]
    \centering
    \begin{subfigure}{0.2\textwidth}
        \centering
        \includegraphics[height=9cm]{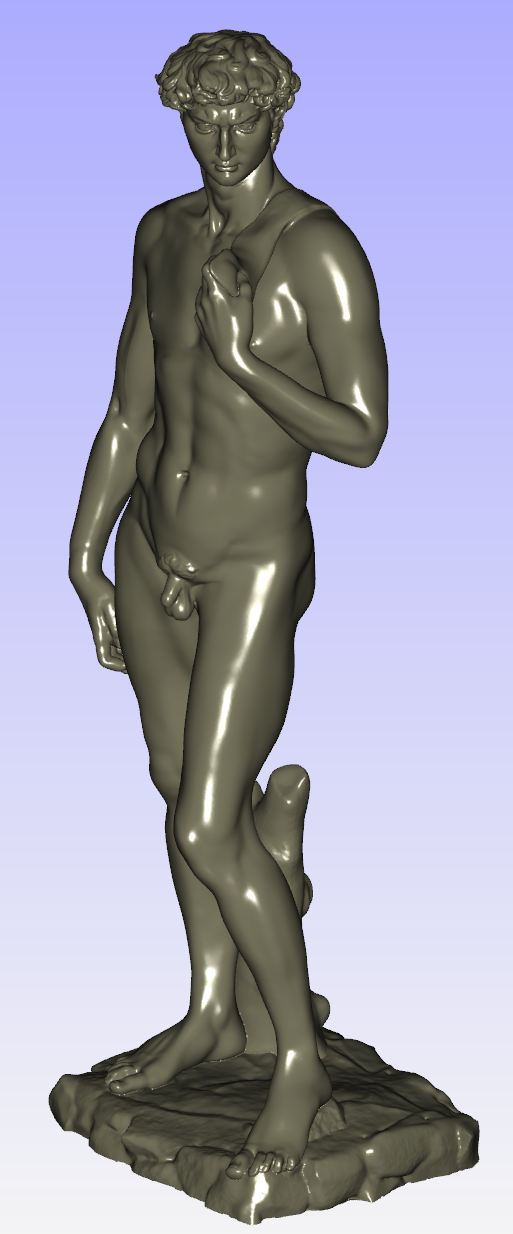}
        \caption{3D mesh model}
        \label{fig:david_mesh}
    \end{subfigure}
    \begin{subfigure}{0.2\textwidth}
        \centering
        \includegraphics[height=9cm]{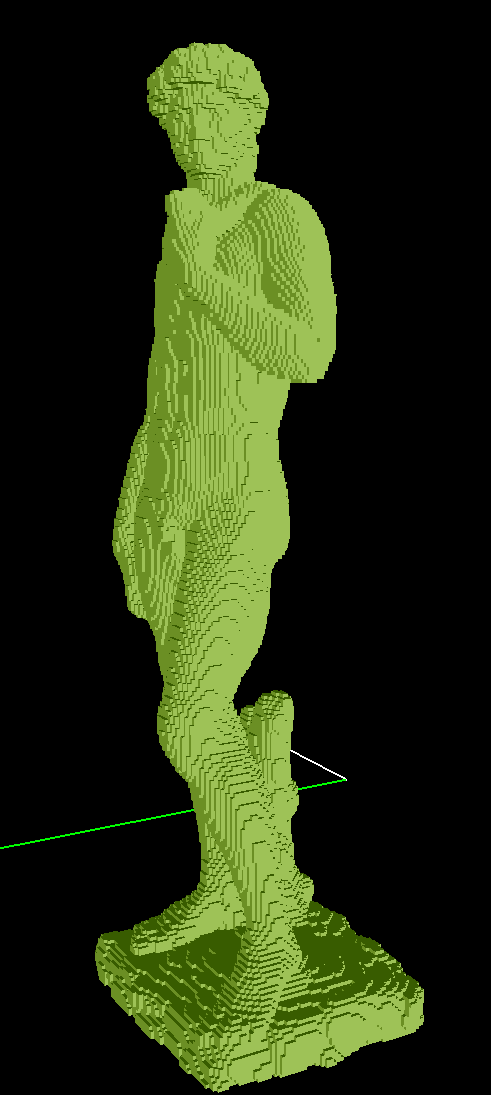}
        \caption{Voxelized model}
        \label{fig:david_vox}
    \end{subfigure}
    \caption{3D model representations of Michelangelo's David}
    \label{fig:david} 
\end{figure}

We first simplified the problem by representing the 3D model as well as the given material as voxels. The primary reason for this was to discretize the workspace, making a search algorithm easily implementable. Further, voxelizing the 3D model allowed us to work at different resolution levels, giving us varying levels of search complexity. Each voxel can have one of two values -- material to be kept and material to be removed. As our robot reaches a voxel with material to be removed, the voxel is removed from the workspace. \Cref{fig:david} shows the voxelization of a 3D mesh model of Michelangelo's Statue of David, which we used for all our experiments.

To establish a baseline for the performance of our search engine, we further simplified the problem by assuming a translating robot instead of a multi-link manipulator. The translating robot can occupy a single voxel at a time and can translate in 6 directions. When the robot moves to a voxel with material to be removed, the voxel is removed from the workspace. With this, we can formulate a simple search algorithm inside the voxel space that will generate a trajectory for the robot that visits every single voxel with material to be removed.

A* search is often used for path planning in voxel space~\cite{brewer2018benchmarks}. It is guaranteed to return an optimal path for a given problem provided that the heuristic used is admissible, i.e., the heuristic underestimates the actual cost to goal from the current node. In searching for a path for sculpting in the voxel space, A* search can be used with a heuristic that estimates the amount of material left to be removed. Each node of the search tree would be a state consisting of the voxel space with current values for each voxel, the location of the robot, and the path of the robot till that point.

While good for general path finding, A* search is not ideal for the purpose of robot sculpting. The  reason is that, since the path has to visit all voxels with material to be removed, it will be extremely long, as a result of which there will be several optimal paths available for the same goal. By design, A* search will only terminate when the guaranteed best path is found, which, with an underestimating heuristic, might not be until all optimal paths have been explored exhaustively. 

Instead of A* search, we use the greedy best-first search. In this algorithm, we ignore the actual cost of the path. Instead, we expand the path with the lowest estimated cost to goal using a heuristic value that may not be admissible. Thus, when covering regions with a large number of similar optimal paths available, only one such path will be expanded. We used greedy best-first search for the translating robot in voxel space with a heuristic that was the sum of the number of voxels left to remove and the $L_1$ distance to the nearest voxel to be removed (\Cref{eq:heuristic_voxel}). The start position of the robot is chosen to be a voxel with material to be removed on an outer face of the voxel space closest to one corner of the voxel space. Separate instances of the search were run for partitioned sections of voxels with material to be removed, i.e.\ different sections of voxels with material to be removed that are completely separated by voxels with material to be kept.
\begin{equation}
    h\left(n\right) = N_{\text{voxels\_left}} + L_1\left(\text{nearest\_voxel}\right)
    \label{eq:heuristic_voxel}
\end{equation}

\begin{figure}
    \centering
    \includegraphics[width=0.4\textwidth]{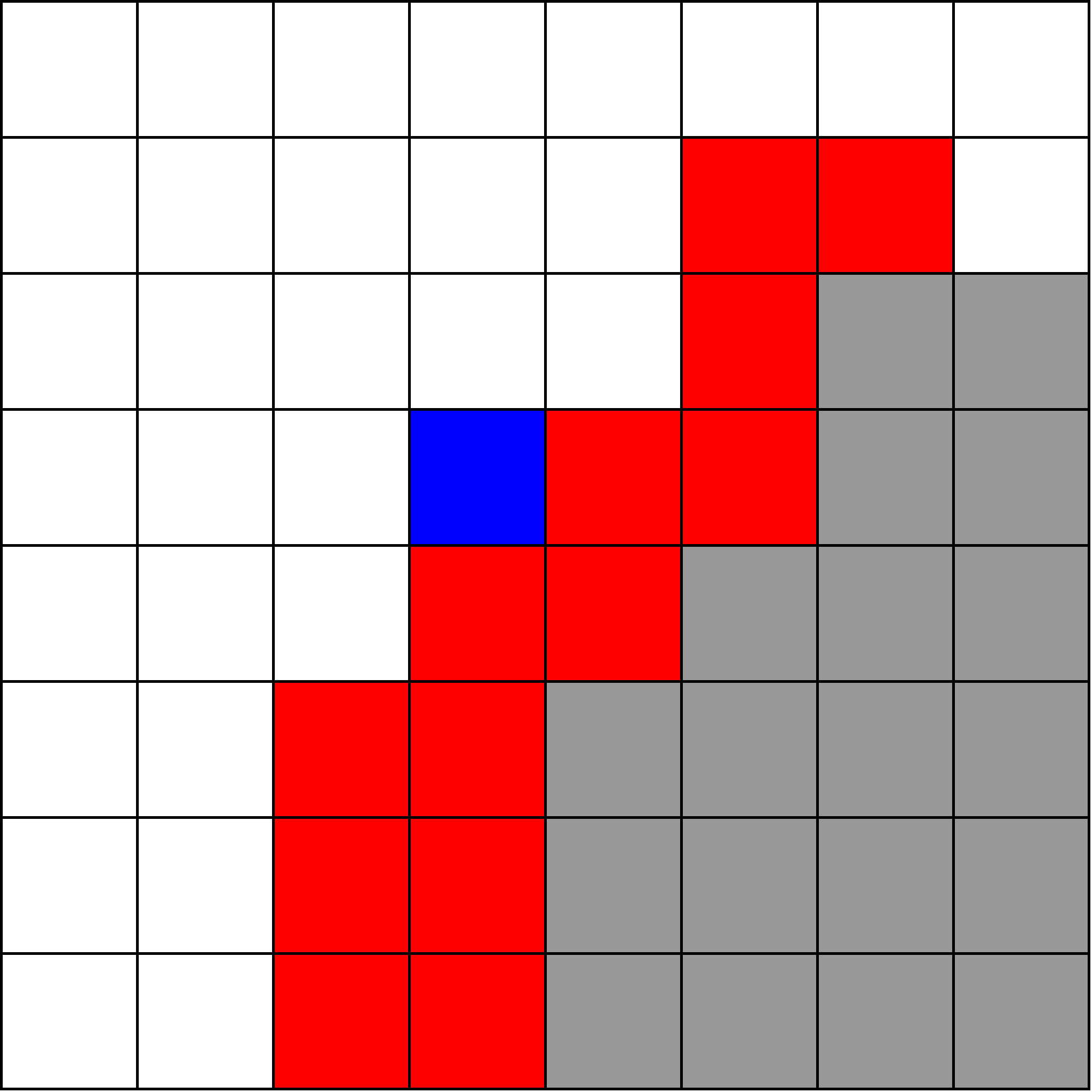}
    \caption{Suboptimal state for A* and best-first search}
    \label{fig:stuck}
\end{figure}

The greedy best-first search algorithm also suffers from some limitations faced by A* search. For instance, in \cref{fig:stuck}, the blue robot, after reaching this node, will not be able to continue to remove the next block with either algorithm, thus forcing the search to begin exhaustively searching previous nodes till it can find a path that leads to the gray voxel with material to remove. With larger sizes of the voxel space, such situations will be hard to avoid even with a well designed heuristic. This would lead to an extremely slow and inefficient search. Thus, the size of the search area must be decreased through some segmentation.


\subsection{Octrees}
\label{sec:octrees}

Octrees are a tree data structure where each non-leaf node has exactly 8 children. A cubic voxel space where each side is of length of a power of 2 can be represented as an octree. Groups of leaf nodes that have a common ancestor and have the same value (i.e.\ material to be removed or material to be kept) can be pruned down to the ancestor node. This can exponentially decrease the number of leaf nodes, thereby creating a significantly smaller search space. It should be noted, however, that the actual search is not carried out in the octree itself, but rather in a graph consisting of leaf nodes of the octree. We will call this graph the \textit{octree-graph}, and each node in this graph a \textit{block}.

\begin{figure*}[h!]
    \centering
    \begin{subfigure}{0.4\textwidth}
        \raggedright
        \includegraphics[width=0.9\linewidth]{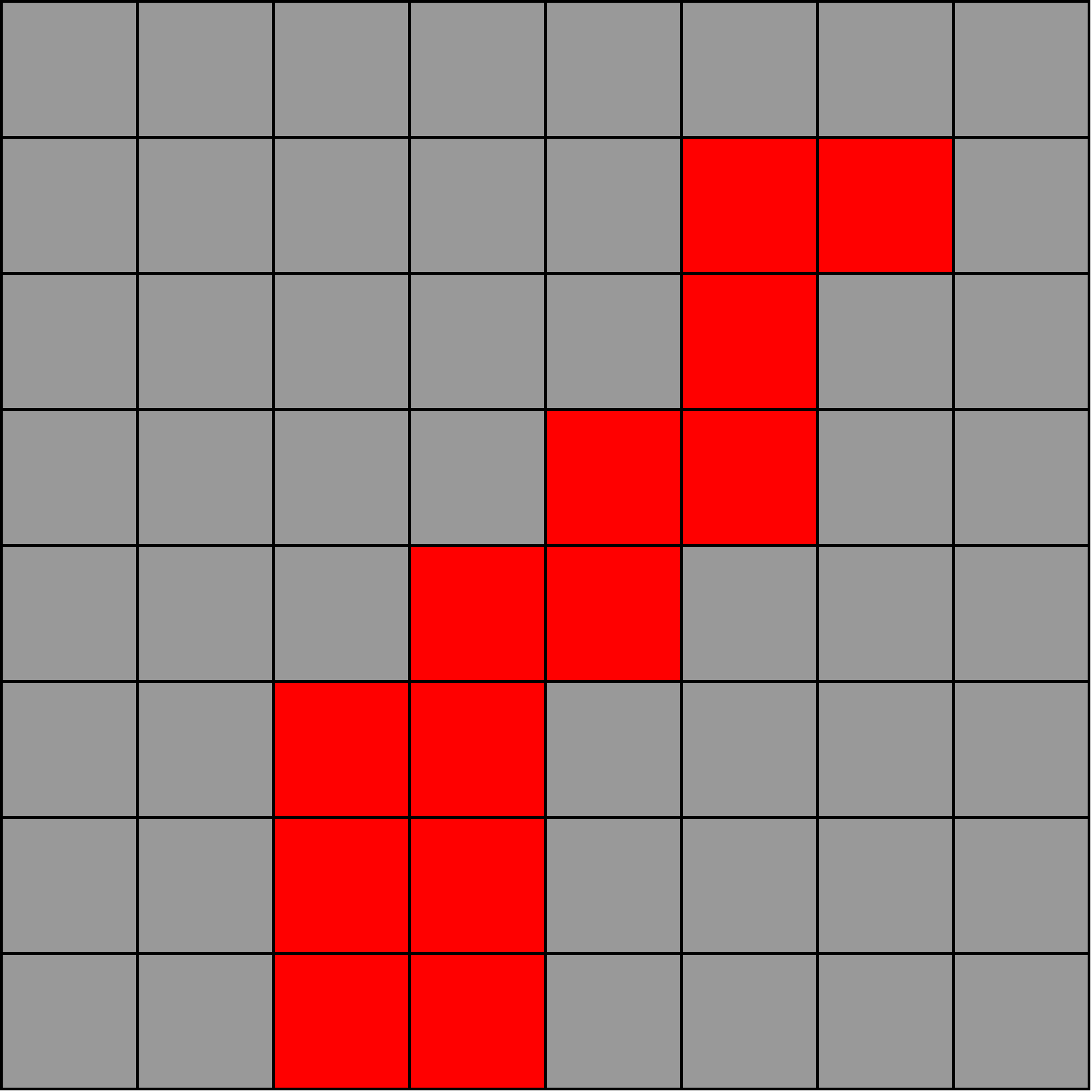}
        \caption{Model in voxel-space}
        \label{fig:quadtree-vox}
    \end{subfigure}
    \begin{subfigure}{0.4\textwidth}
        \raggedleft
        \includegraphics[width=0.9\linewidth]{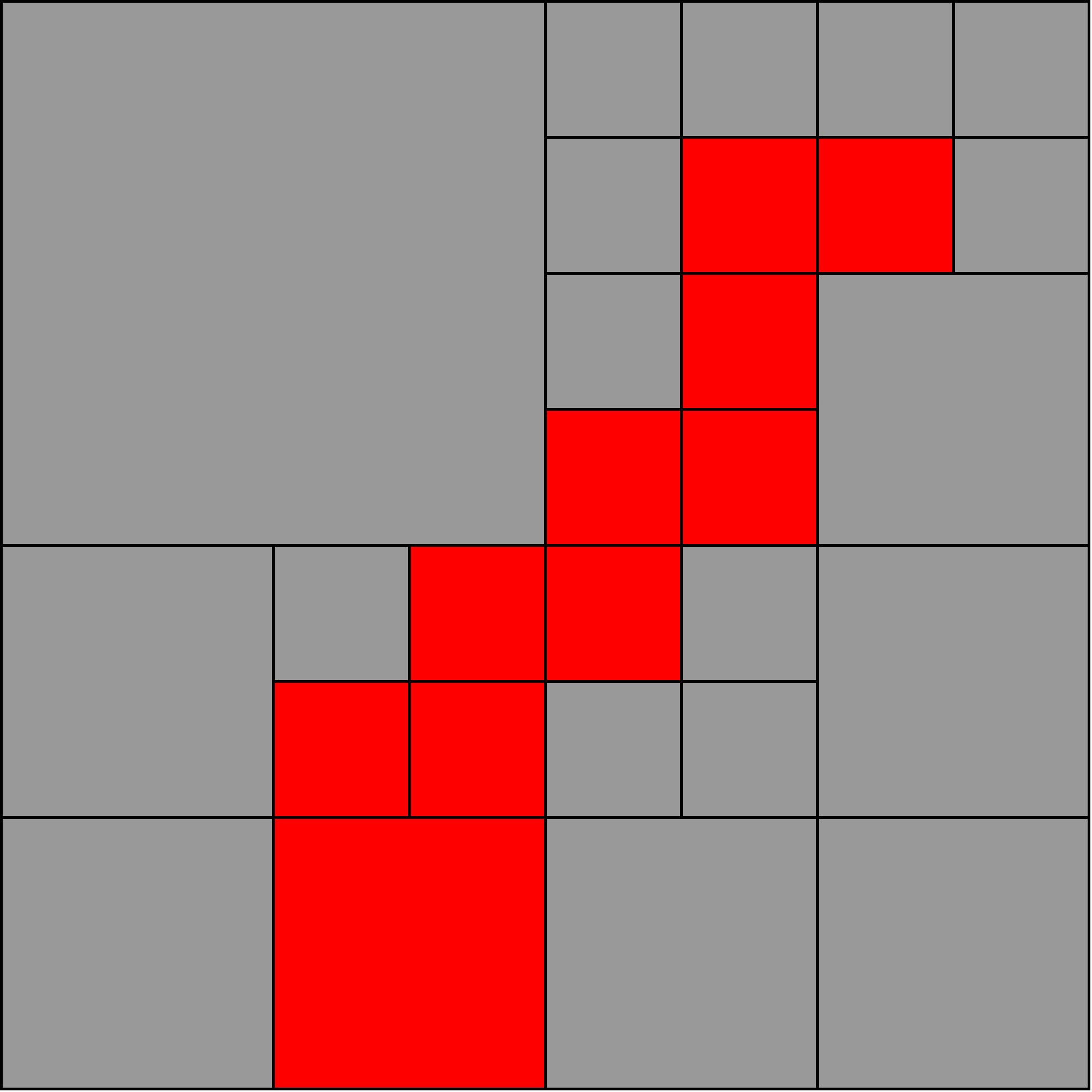}
        \caption{Model in quadtree-graph}
        \label{fig:quadtree_full}
    \end{subfigure}
    \caption{Voxel to quadtree-graph conversion}
    \label{fig:quadtree} 
\end{figure*}

A voxel graph is turned into an orders of magnitude smaller octree-graph. \Cref{fig:quadtree} shows the conversion of a voxel grid to a quadtree-graph. The concept is similar to that of octrees, but in 2D instead of 3D. \Cref{fig:quadtree-vox} shows the 2D pixel representation of a simple figure, with red pixels representing material to keep and gray pixels representing material to remove. The quadtree-graph in \cref{fig:quadtree_full} shows gray blocks with material to remove and red blocks with material to keep. As can be seen, the pixel graph has 64 pixels, while the quadtree-graph has just 28 blocks, a reduction of $56\%$. Larger voxel graphs see a larger decrease in the number of nodes. Results are shown in \Cref{sec:results} and in \Cref{table:octree_size}.

    Greedy best-first search with the same heurisitc as \Cref{eq:heuristic_voxel} can be used in the octree space. The heuristic favors removing larger blocks. The $L_1$ distance is computed in the voxel space. The branching factor of the search is higher since a block in the octree-graph can have more than one neighbor on each side, resulting in more than six adjacent blocks. These neighbors can be computed using an algorithm from~\cite{Samet1989}. However, a higher branching factor does not increase the time complexity of the best-first search.

With the decomposition of the voxel space into octree blocks, a boustrophedon trajectory inside each block for the translating robot~\cite{Choset2000} can be used. Those boustrophedon trajectories can be joined together using simple trajectories on a single plane since those paths are only along an edge of a block. Traversal in empty blocks is also done in a similar manner, by moving along each direction once till the destination coordinates are reached. While using naive trajectories seems to defeat the purpose of the search based motion planning proposed in this paper, it is still essential for the multi-link manipulator. The naive trajectories are possible for the translating robot simply because collision avoidance in the workspace is satisfactory for it.

\subsection{Multi-Link Manipulator}
\label{sec:arm}

Several new challenges are introduced when replacing the translating robot with a multi-link manipulator. As mentioned in \Cref{sec:octrees}, we can no longer use a naively generated path inside the octree blocks. Instead, we must use another search based method to generate those collision free paths. Further, in our previous approach, the translating robot had a discrete state where it occupied a single voxel. This is no longer the case with a multi-link manipulator.

\begin{figure}[h]
    \centering
    \begin{subfigure}[t]{0.2\textwidth}
        \raggedright
        \includegraphics[width=0.9\linewidth]{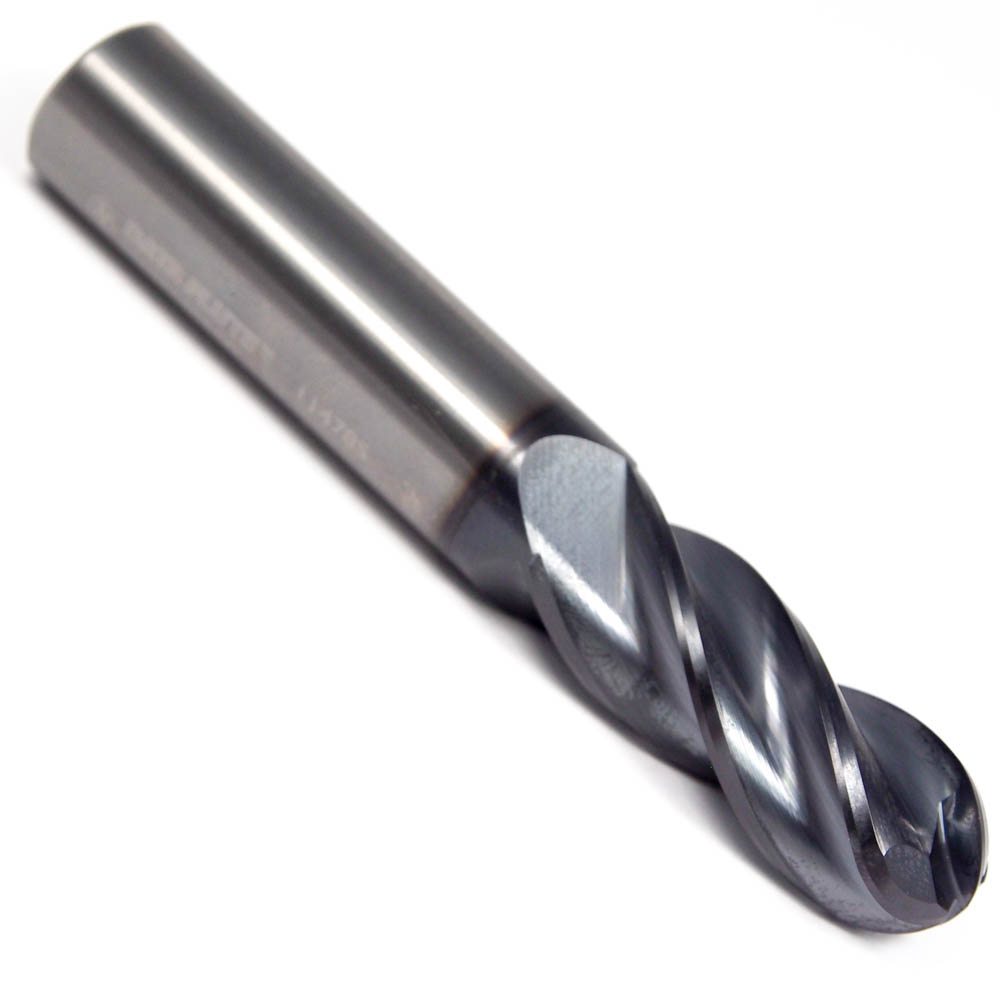}
        \caption{Ball End Mill}
    \end{subfigure}
    \begin{subfigure}[t]{0.2\textwidth}
        \raggedleft
        \includegraphics[width=0.9\linewidth]{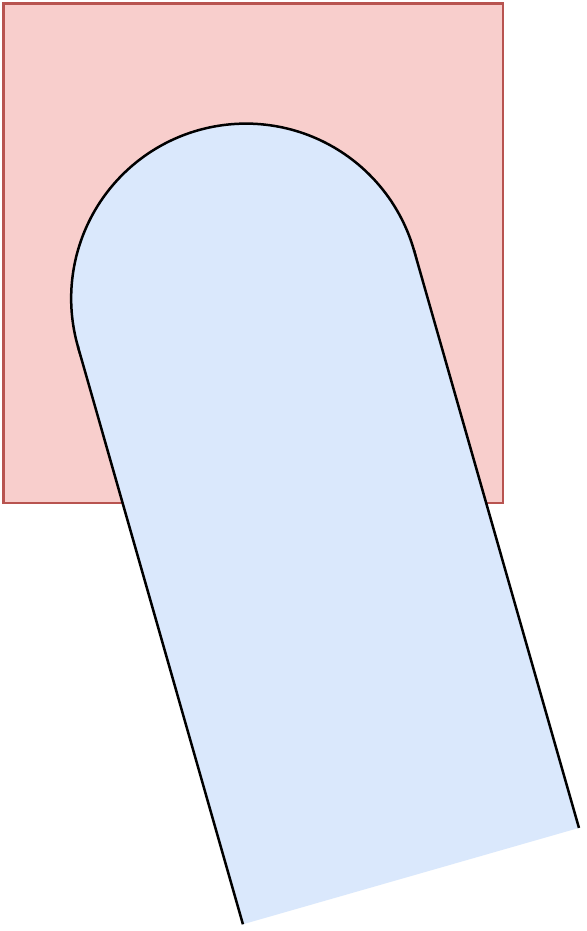}
        \caption{Virtual Ball End Mill inside a Voxel}
    \end{subfigure}
    \caption{The ball-end mill fits perfectly inside a voxel}
    \label{fig:ballend_voxel} 
\end{figure}

The problem of removing material in discrete voxels with a robot end effector in continuous space can be solved by establishing a criterion for satisfactory removal of an entire voxel. As we described in \Cref{sec:voxel}, each voxel can be considered to be roughly the size of the tip of a ball-end mill. We define reaching the center of a voxel with the ball-end of the mill without colliding with any other voxels or blocks to be sufficient to completely remove that voxel. This can be seen in \Cref{fig:ballend_voxel}.

The broader search algorithm in the octree-graph remains the same in implementation but has an added constraint -- not all neighboring octree blocks of the current octree block may be removable since a collision free removal trajectory inside those blocks may not be possible with the current state of the voxel space. Thus, we must check whether a block is entire removable before expanding the search to that node.

A second greedy best-first search is required inside each block in order to determine whether the block is removable as well as to generate a trajectory for the manipulator to remove that block. The search begins with the robot at a voxel on the approach face of the block. At each node of the search tree, the tree can expand to any voxel inside the block that has at least one face open. Expanding to a voxel involves attempting a trajectory from the current voxel to that voxel. If a trajectory is not possible, that voxel is not considered. If no trajectories are possible to any of the voxels with open faces within the block, the block is unremovable at that state, and is removed from that state of the outer search. A postprocessing step removes all extra trajectories that do not reach a voxel to remove material and replans intermittent trajectories as necessary. The motion planning for the search is done using the Open Mation Planning Library (OMPL)~\cite{sucan2012open} and collision avoidance is implemented using the Flexible Collision Library~\cite{pan2012fcl}. A URDF model of the robot is used to check for collisions with both the remaining octree blocks as well as the individual voxels inside the current block with all parts of the robot.

The heuristic used favors voxels that are closest to the current voxel. Linear trajectories are also preferred. Voxels that are not in-line with the previous two voxels removed have a higher heuristic value in order to encourage more linear paths. Finally milling "inwards" is also not preferred. The heuristic value is increased for directions opposing any of the open faces of the block.

\section{Experiments and Results}
\label{sec:results}

We evaluated all three algorithms in simulation using a 3D model of Michelangelo's Statue of David as shown in \Cref{fig:david}. In particular, we tested the pure voxel space search with the translating robot, the octree-based method with the translating robot, and the octree-based method with the multi-link manipulator. 
We voxelized the 3D mesh model at different voxel resolutions using \emph{binvox}~\cite{min2004binvox}, which uses the methods described in~\cite{nooruddin2003simplification} to generate the voxel model.

\begin{table}[ht]
    \begin{center}
    \renewcommand{\arraystretch}{1.2}
    \begin{tabular}{l|l|l}
    Voxel Resolution & Voxel Representation & Octree Representation\\
    \hline
    $8 \times 8 \times 8$ & $12$ & $14$\\
    $16 \times 16 \times 16$ & $65$ & $25$\\
    $64 \times 64 \times 64$ & $2745$ & $209$\\
    $256 \times 256 \times 256$ & N/A & $1265$\\
    \end{tabular}
    \end{center}
    
    \caption{Search time (seconds) for translating robot, respectively with voxel and octree representations.}
    \label{table:translating_time}
    \vspace{-0.2cm}
\end{table}

\subsection{Voxel Approach}

The pure voxel space search with the translating robot becomes impractical very quickly. We tested the basic voxel space search approach on 4 different voxel resolutions of the 3D model. As can be seen in \Cref{table:translating_time}, the search time grows substantially from $8 \times 8 \times 8$ resolution to $64 \times 64 \times 64$. The search time is impractical at resolutions beyond that.

\subsection{Octree}

\begin{figure}[h]
    \centering
    \includegraphics[width=0.49\textwidth]{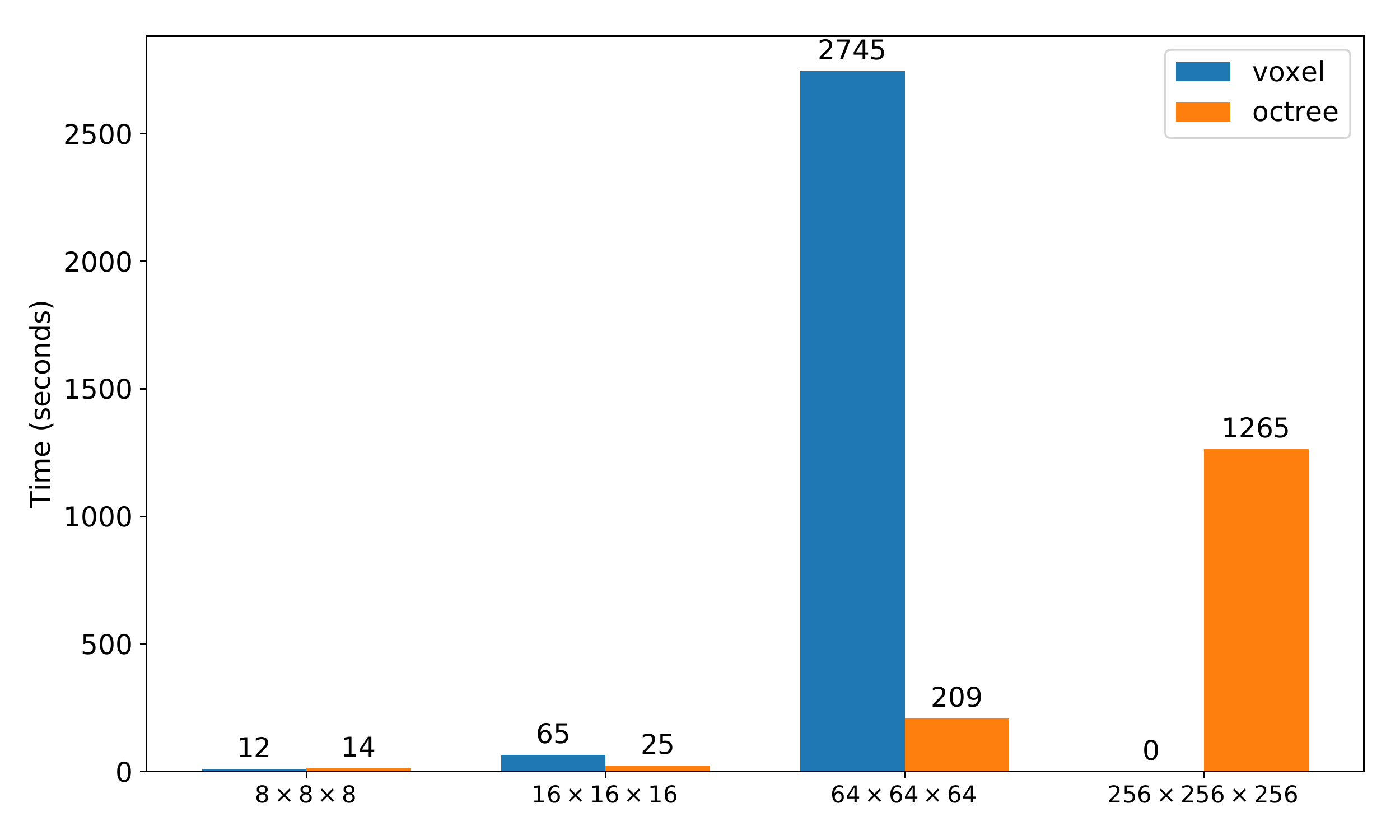}
    \caption{Time for completion for search in voxels and octrees for different voxel resolutions with the translating robot}
    \label{fig:time_chart}
\end{figure}

Using the octree representation, the complexity of the search is much reduced.
We performed similar tests for the octree approach with the translating robot. 
We measured the time taken for the search on the same four voxel resolutions. The trajectories inside each block were computed naively. \Cref{table:translating_time} and \Cref{fig:time_chart} show the search time in comparison to the non-octree approach. 

\begin{table}[h]
    \begin{center}
    \renewcommand{\arraystretch}{1.2}
    \begin{tabular}{l|l|l|l}
    Voxel Resolution & Voxel Count & Octree Count & \% Decrease\\
    \hline
    $8 \times 8 \times 8$ & $512$ & $354$ & $30.9\%$ \\
    $16 \times 16 \times 16$ & $4096$ & $1756$ & $57.1\%$\\
    $64 \times 64 \times 64$ & $262144$ & $53512$ & $79.6\%$\\
    $256 \times 256 \times 256$ & $16777216$ & $793403$ & $95.3\%$
    \end{tabular}
    \end{center}
    
    \caption{Voxel and octree graph sizes}
    \label{table:octree_size}
    \vspace{-0.2cm}
\end{table}

The decrease in the size of the search space leads to heavy speedup of the search algorithm on octrees. To better quantify this, we evaluated the percent decrease in the number of nodes from the voxel graph to the octree graph at all resolutions. As can be seen in \Cref{table:octree_size}, larger and larger percentage of voxels are grouped up into blocks in the octree conversion. In fact, the $256 \times 256 \times 256$ voxel resolution in the octree representation has just a few times more blocks than the number of voxels in the $64 \times 64 \times 64$ voxel resolution without the octree representation.

\subsection{Multi-Axis Manipulator}

We tested the third algorithm, which takes into account collisions with the manipulator, using a simulation of a Franka Emika Panda robot in a ROS Gazebo environment. This is a 7-link redundant manipulator with high dexterity. Due to the circular workspace of the manipulator, we added an additional rotational axis through the center of the voxel grid, allowing the robot to access all sides of the material voxel grid with ease. The virtual mill was modeled simply as a cylinder with a spherical end. The size of the sphere was slightly smaller than the size of the voxels, as shown in \Cref{fig:ballend_voxel}. This was done in order to allow the robot some freedom in orientation when removing a voxel accessible from a single face rather than forcing it to approach the voxel perfectly perpendicular to an open face.

\begin{figure*}[!ht]
    \centering
    \begin{subfigure}{0.49\textwidth}
        \includegraphics[height=6cm]{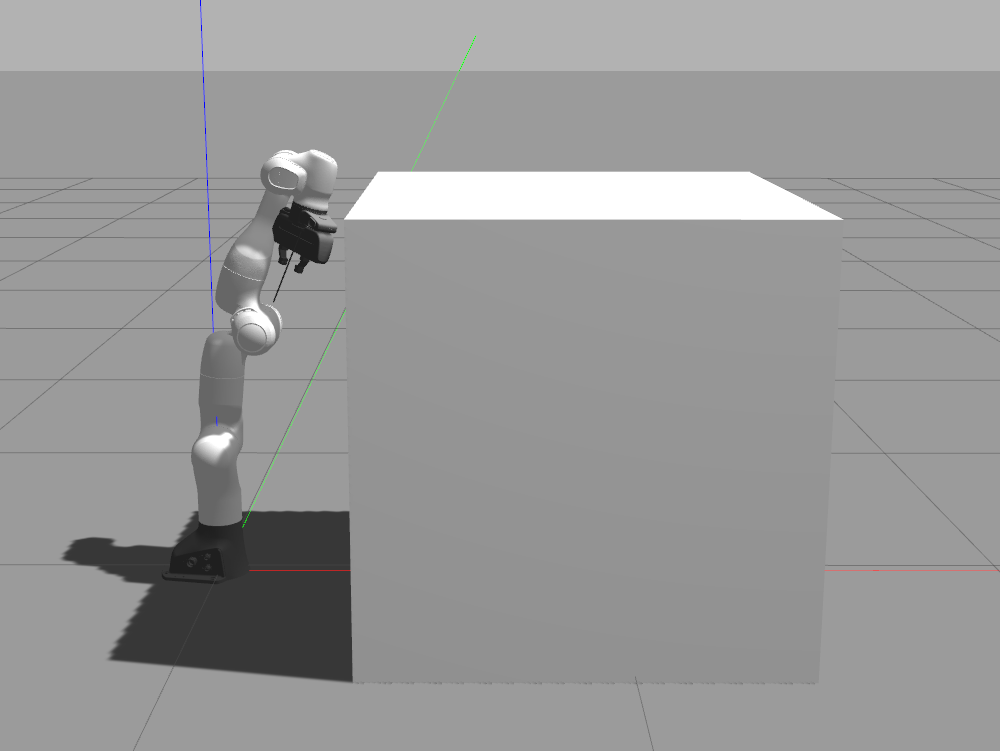}
        \caption{Before Sculpting}
        \label{fig:sim_full}
    \end{subfigure}
    \begin{subfigure}{0.49\textwidth}
        \includegraphics[height=6cm]{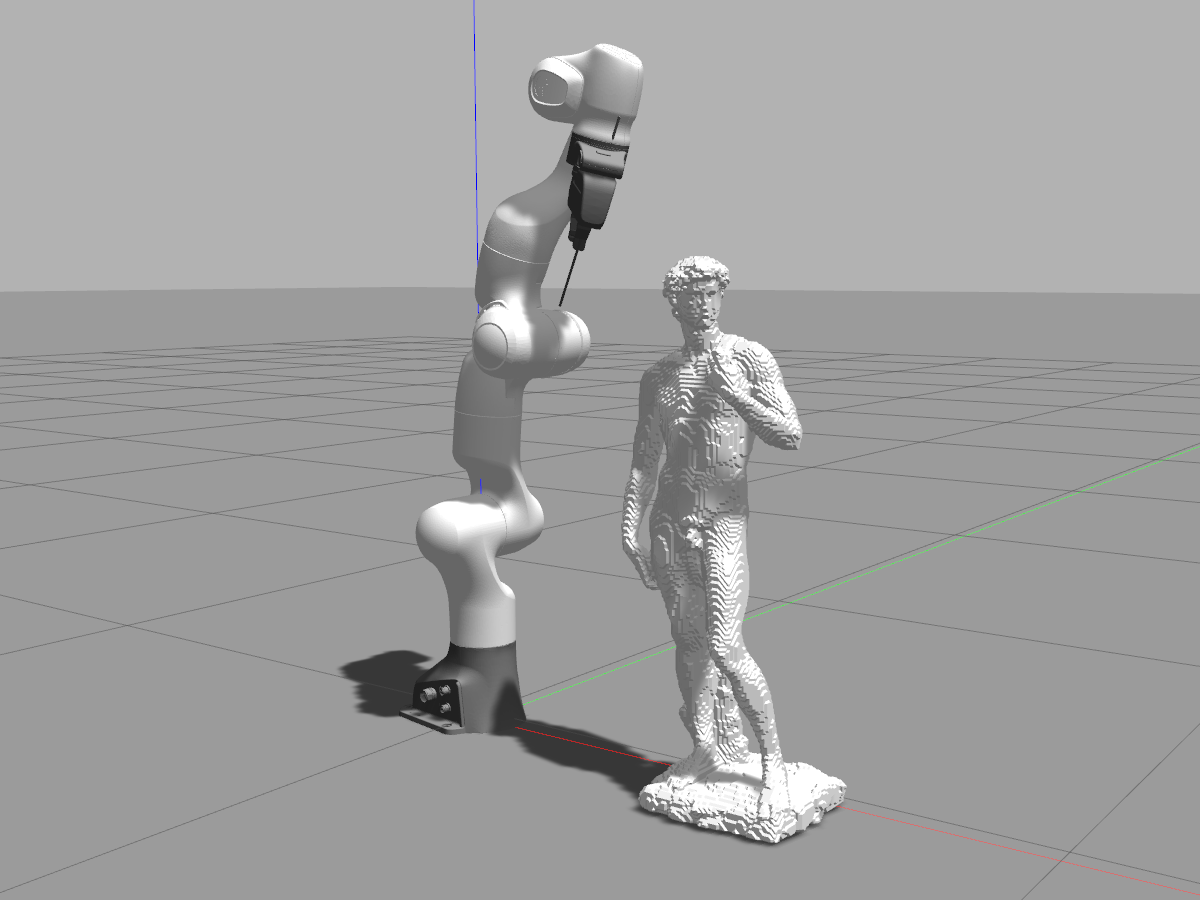}
        \caption{$256 \times 256 \times 256$ voxel model after sculpting}
        \label{fig:sim_256}
    \end{subfigure}
    \caption{Simulation environment with the Franka robot before and after sculpting}
    \label{fig:arm_sculptures} 
\end{figure*}

The algorithm generates complete trajectories for all voxel resolutions of the 3D model and is able to sculpt the statue successfully without any collisions. \Cref{fig:arm_sculptures} shows the simulation environment with different voxel resolutions of the 3D model.


\begin{table}
    \begin{center}
    \renewcommand{\arraystretch}{1.2}
    \begin{tabular}{l|l}
    Voxel Resolution & Manipulator Time (minutes) \\
    \hline
    $16 \times 16 \times 16$ & $7$\\
    $64 \times 64 \times 64$ & $47$\\
    $256 \times 256 \times 256$ & $457$
    \end{tabular}
    \end{center}
    
    \caption{Search time for multi-link manipulator}
    \label{table:arm_time}
    \vspace{-0.2cm}
\end{table}

However, as expected, the search for a feasible trajectory takes significantly more time than in the case for a translating robot, even with the octree representation. \Cref{table:arm_time} shows the time taken for generating trajectories of different voxel sizes. We only considered the time taken for the actual motion planning, not the time spent on execution of the plans. Running these complex motion plans can take several hours while doing this for the translating robot is instantaneous.
The much larger search times are explained by the collision checking inside each block. In particular, each branch in the search tree requires the generation of a motion plan with an external library. 

\begin{figure}
    \centering
    \includegraphics[width=0.4\textwidth]{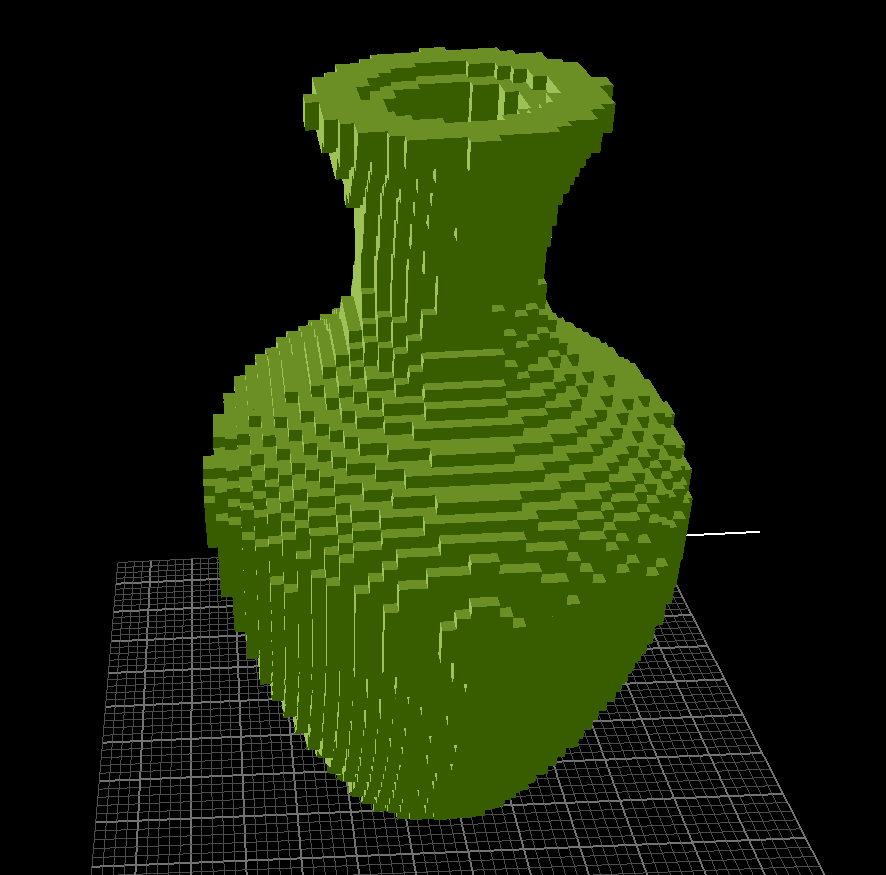}
    \caption{The robot cannot sculpt this vase since the inner voxels are unreachable}
    \label{fig:vase}
\end{figure}

We also tested the algorithm in a failure scenario where the robot is unable to complete the search due to inaccessible blocks. After removing the outer material around the hollow vase in \Cref{fig:vase}, the robot was unable to remove the voxels inside the vase through the neck. Thus, the search terminated at that point.

\section{Future Work}
\label{sec:future}

With our proposed solution, we were successfully able to sculpt the Statue of David out of voxels in simulation with a robotic arm. However, our approach does have several limitations that we would like to address in the future.

\subsection{Experimental Validation}

Due to hardware limitations, we were unable to test our algorithms on a physical robot. There are several challenges that we anticipate when running the sculpting process on a physical robot. First, due to our approximation of voxel size with respect to the size of the ball-end mill, the motion plans may not remove the entire material of a voxel. While this will not be an issue for most of the material removal, it will create course patterns on the final surface finish of the sculpture. The surface finish's coarseness will also vary depending on the linearity of the voxel space trajectories along the surface -- more linear trajectories will create a smoother surface but trajectories that often diverge will create more coarse patterns.

\subsection{Expanding to Freeform Surfaces}
\label{sec:freeform}

Several of the limitations mentioned above arise due to the use of voxels for discretizing the search space. These problems can be solved using freeforms surface to represent the 3D models. This is a more challenging task since there is no obvious method of segmenting the models into discrete, searchable components. A solution considered prior to the voxel approach involved finding all patches on the surface of the 3D model that the robot can cover without any collisions. Coverage of each patch results in a morphological erosion of the material by a sphere defined by the ball-end mill of the robot. With these discrete decisions, our search algorithm can be used where the patches are analogous to blocks in the octree-graph. New patches will be generated after removal of each patch. The primary challenge with this solution is the search for the patches on the surface of the material. While feasible in 2D, where the search can be carried in one of two directions of the curve, a surface in 3D will require a search with infinite directions.

\subsection{3D Visualization}

So far we have been using simple plotting libraries to visualize the voxel approach. While this is adequate for small voxel-spaces, a more robust solution is necessary for larger models. For the multi-link manipulator experiments, we visualized the process in Gazebo. However, higher voxel resolutions tended to slow down the simulation. We would like to use an advanced 3D library or software to visualize each state of the model with high quality detail and lighting. We may also expand this idea to a complete sculpting process visualization in 3D and perhaps in Augmented Reality (AR). This would provide a straightforward way of showcasing the sculpting process without wasting material and energy.

\section{Conclusion}

In this paper, we have proposed an algorithm for generating complete collision-free trajectories for material removal for sculpting with a robotic manipulator. Our solution is completely generalized and works as expected for voxel representations of 3D models. We have evaluated it by testing it in simulation. We have shown the time consumed for generating the paths. We have also discussed the possible future works for this project.

\clearpage

{\small
\bibliographystyle{IEEEtran}
\bibliography{bibliography}

\begin{thebibliography}{10}
\providecommand{\url}[1]{#1}
\csname url@rmstyle\endcsname
\providecommand{\newblock}{\relax}
\providecommand{\bibinfo}[2]{#2}
\providecommand\BIBentrySTDinterwordspacing{\spaceskip=0pt\relax}
\providecommand\BIBentryALTinterwordstretchfactor{4}
\providecommand\BIBentryALTinterwordspacing{\spaceskip=\fontdimen2\font plus
\BIBentryALTinterwordstretchfactor\fontdimen3\font minus
  \fontdimen4\font\relax}
\providecommand\BIBforeignlanguage[2]{{%
\expandafter\ifx\csname l@#1\endcsname\relax
\typeout{** WARNING: IEEEtran.bst: No hyphenation pattern has been}%
\typeout{** loaded for the language `#1'. Using the pattern for}%
\typeout{** the default language instead.}%
\else
\language=\csname l@#1\endcsname
\fi
#2}}

\bibitem{xuejuan2007robot}
N.~Xuejuan, L.~Jingtai, S.~Lei, L.~Zheng, and C.~Xinwei, ``Robot 3d sculpturing
  based on extracted nurbs,'' in \emph{2007 IEEE International Conference on
  Robotics and Biomimetics (ROBIO)}.\hskip 1em plus 0.5em minus 0.4em\relax
  IEEE, 2007, pp. 1936--1941.

\bibitem{lei20083d}
S.~Lei, C.~Xinwei, L.~Jingtai, L.~Zheng, and N.~Xuejuan, ``3d terrain model
  approach by an industrial robot,'' in \emph{2008 7th World Congress on
  Intelligent Control and Automation}.\hskip 1em plus 0.5em minus 0.4em\relax
  IEEE, 2008, pp. 2345--2349.

\bibitem{lasemi2010recent}
A.~Lasemi, D.~Xue, and P.~Gu, ``Recent development in cnc machining of freeform
  surfaces: a state-of-the-art review,'' \emph{Computer-Aided Design}, vol.~42,
  no.~7, pp. 641--654, 2010.

\bibitem{ilushin2005precise}
O.~Ilushin, G.~Elber, D.~Halperin, R.~Wein, and M.-S. Kim, ``Precise global
  collision detection in multi-axis nc-machining,'' \emph{Computer-Aided
  Design}, vol.~37, no.~9, pp. 909--920, 2005.

\bibitem{jun2003optimizing}
C.-S. Jun, K.~Cha, and Y.-S. Lee, ``Optimizing tool orientations for 5-axis
  machining by configuration-space search method,'' \emph{Computer-Aided
  Design}, vol.~35, no.~6, pp. 549--566, 2003.

\bibitem{Choi1997}
B.~K. Choi, D.~H. Kim, and R.~B. Jerard, ``{C-space approach to tool-path
  generation for die and mould machining},'' \emph{CAD Computer Aided Design},
  vol.~29, no.~9, pp. 657--669, 1997.

\bibitem{Morishige1997}
K.~Morishige, K.~Kase, and Y.~Takeuchi, ``{Collision-free tool path generation
  using 2-dimensional C-space for 5-axis control machining},''
  \emph{International Journal of Advanced Manufacturing Technology}, vol.~13,
  no.~6, pp. 393--400, 1997.

\bibitem{chen2005local}
T.~Chen, P.~Ye, and J.~Wang, ``Local interference detection and avoidance in
  five-axis nc machining of sculptured surfaces,'' \emph{The International
  Journal of Advanced Manufacturing Technology}, vol.~25, no. 3-4, pp.
  343--349, 2005.

\bibitem{Bo2016}
\BIBentryALTinterwordspacing
P.~Bo, M.~Bartoň, D.~Plakhotnik, and H.~Pottmann, ``{Towards efficient 5-axis
  flank CNC machining of free-form surfaces via fitting envelopes of surfaces
  of revolution},'' \emph{CAD Computer Aided Design}, vol.~79, pp. 1--11, 2016.
  [Online]. Available: \url{http://dx.doi.org/10.1016/j.cad.2016.04.004}
\BIBentrySTDinterwordspacing

\bibitem{Wang2006}
Q.~H. Wang, J.~R. Li, and R.~R. Zhou, ``{Graphics-assisted approach to rapid
  collision detection for multi-axis machining},'' \emph{International Journal
  of Advanced Manufacturing Technology}, vol.~30, no. 9-10, pp. 853--863, 2006.

\bibitem{Lauwers2003a}
B.~Lauwers, P.~Dejonghe, and J.~P. Kruth, ``{Optimal and collision free tool
  posture in five-axis machining through the tight integration of tool path
  generation and machine simulation},'' \emph{CAD Computer Aided Design},
  vol.~35, no.~5, pp. 421--432, 2003.

\bibitem{tang2014algorithms}
T.~D. Tang, ``Algorithms for collision detection and avoidance for five-axis nc
  machining: a state of the art review,'' \emph{Computer-Aided Design},
  vol.~51, pp. 1--17, 2014.

\bibitem{Choset2001}
H.~Choset, ``{Coverage for robotics – A survey of recent results},''
  \emph{Annals of Mathematics and Artificial Intelligence}, vol.~31, no. 1-4,
  pp. 113--126, 2001.

\bibitem{Choset2000}
------, ``{Coverage of known spaces: The boustrophedon cellular
  decomposition},'' in \emph{Autonomous Robots}, vol.~9, no.~3, 2000, pp.
  247--253.

\bibitem{Atkar2005}
P.~N. Atkar, A.~Greenfield, D.~C. Conner, H.~Choset, and A.~A. Rizzi,
  ``{Uniform coverage of automotive surface patches},'' \emph{International
  Journal of Robotics Research}, vol.~24, no.~11, pp. 883--898, 2005.

\bibitem{breitenmoser2010distributed}
A.~Breitenmoser, J.-C. Metzger, R.~Siegwart, and D.~Rus, ``Distributed coverage
  control on surfaces in 3d space,'' in \emph{2010 IEEE/RSJ International
  Conference on Intelligent Robots and Systems}.\hskip 1em plus 0.5em minus
  0.4em\relax IEEE, 2010, pp. 5569--5576.

\bibitem{zaplana2018novel}
I.~Zaplana and L.~Basanez, ``A novel closed-form solution for the inverse
  kinematics of redundant manipulators through workspace analysis,''
  \emph{Mechanism and Machine Theory}, vol. 121, pp. 829--843, 2018.

\bibitem{Hess2012}
J.~Hess, G.~D. Tipaldi, and W.~Burgard, ``{Null space optimization for
  effective coverage of 3D surfaces using redundant manipulators},'' \emph{IEEE
  International Conference on Intelligent Robots and Systems}, pp. 1923--1928,
  2012.

\bibitem{brewer2018benchmarks}
D.~Brewer and N.~R. Sturtevant, ``Benchmarks for pathfinding in 3d voxel
  space,'' in \emph{Eleventh Annual Symposium on Combinatorial Search}, 2018.

\bibitem{Samet1989}
H.~Samet, ``{Neighbor finding in images represented by octrees},''
  \emph{Computer Vision, Graphics and Image Processing}, vol.~46, no.~3, pp.
  367--386, 1989.

\bibitem{sucan2012open}
I.~A. Sucan, M.~Moll, and L.~E. Kavraki, ``The open motion planning library,''
  \emph{IEEE Robotics \& Automation Magazine}, vol.~19, no.~4, pp. 72--82,
  2012.

\bibitem{pan2012fcl}
J.~Pan, S.~Chitta, and D.~Manocha, ``Fcl: A general purpose library for
  collision and proximity queries,'' in \emph{2012 IEEE International
  Conference on Robotics and Automation}.\hskip 1em plus 0.5em minus
  0.4em\relax IEEE, 2012, pp. 3859--3866.

\bibitem{min2004binvox}
P.~Min, ``Binvox 3d mesh voxelizer,'' \emph{Available on: http://www. cs.
  princeton. edu/\~{} min/binvox}, 2004.

\bibitem{nooruddin2003simplification}
F.~S. Nooruddin and G.~Turk, ``Simplification and repair of polygonal models
  using volumetric techniques,'' \emph{IEEE Transactions on Visualization and
  Computer Graphics}, vol.~9, no.~2, pp. 191--205, 2003.

\end{thebibliography}
}

\end{document}